\documentclass[a4paper]{article}
\pdfoutput=1
\usepackage{INTERSPEECH2021}
\usepackage{latexsym}
\usepackage{amsmath}
\usepackage{url}
\usepackage{amssymb}
\usepackage{amsfonts}
\usepackage{graphicx}
\usepackage{tabularx}
\usepackage{multirow}
\usepackage{arydshln}
\usepackage{mathtools,nccmath}
\usepackage[T5]{fontenc}
\usepackage{enumitem}
\usepackage{todonotes}
\usepackage{cuted}
\usepackage[most]{tcolorbox}
\usepackage{microtype}
\usepackage[hidelinks]{hyperref}

\title{Intent Detection and Slot Filling for Vietnamese}
\name{Mai Hoang Dao$^\ast$, Thinh Hung Truong\sthanks{\ \ The first two authors contributed equally to this work.} , Dat Quoc Nguyen}
\address{VinAI Research, Hanoi, Vietnam}
\email{\{v.maidh3, v.thinhth88, v.datnq9\}@vinai.io}

\begin{document}

\maketitle
\begin{abstract}
Intent detection and slot filling are important tasks in spoken and natural language understanding. However, Vietnamese is a low-resource language in these research topics. In this paper, we present the \emph{first} public intent detection and slot filling dataset for Vietnamese. In addition, we also propose a joint model  for intent detection and slot filling, that extends the recent state-of-the-art JointBERT+CRF model \cite{abs-1902-10909} with an intent-slot attention layer to explicitly incorporate intent context information  into slot filling via ``soft'' intent label embedding. Experimental results on our  Vietnamese dataset show that our proposed model significantly outperforms JointBERT+CRF. We publicly release our dataset and the implementation of our model at: \url{https://github.com/VinAIResearch/JointIDSF}.
\end{abstract}


\section{Introduction}

Spoken language understanding (SLU) is a crucial component of task-oriented dialogue systems, that typically handles  natural language understanding tasks including intent detection and slot filling. In particular, intent detection aims to identify speaker's intent from a given utterance, while slot filling is to extract from the utterance the correct argument value for the slots of the intent \cite{louvan2020recent}. Despite being the 17th most spoken language in the world \cite{Ethnologue} (about 100M speakers), data resources for Vietnamese SLU are limited. There is only one Vietnamese dataset relevant to intent detection, which is a dialog act corpus containing ISO-24617-2 based annotations over communication acts \cite{NGO18.942}, however, this corpus is not publicly available for the research community. 
To the best of our knowledge, there is no public Vietnamese dataset available specifically for either intent detection or slot filling.

In this paper, we present the {first} public dataset for Vietnamese intent detection and slot filling. We create this dataset through three manual phases. The first phase manually translates each English utterance from the well-known intent detection and slot filling  dataset ATIS \cite{price-1990-evaluation} into Vietnamese. Note that this is not a direct translation as performed for the multilingual ATIS datasets \cite{upadhyay2018almost,xu2020end} where American-specific entities in English are kept intact during translation. In fact, we require modifications to ensure that our Vietnamese utterances are natural, fitting in real-world scenarios in Vietnam and of high-quality, e.g. replacing American-popular slot values, such as locations, airline names and the like, with their  counterparts in Vietnam. In the second manual phase, we project intent and slot annotations from each ATIS English utterance to its Vietnamese-translated version. In the last phase, we manually fix inconsistencies among projected annotations in our Vietnamese dataset.

Recent research on intent detection and slot filling have shown that jointly learning these two tasks helps improve performance results   \cite{louvan2020recent,zhang-etal-2019-joint,Weld2021ASO}. In addition, previous works \cite{goo2018slot,li-etal-2018-self,e-etal-2019-novel,qin-etal-2019-stack,zhang2019joint} present attention mechanisms to incorporate intent context information via an utterance representation into slot filling to boost the performances. We argue that instead of using the utterance representation, we can  incorporate more explicit intent context information via a ``soft'' intent label embedding that is computed based on intent prediction probabilities. Thus, we present a new joint model for intent detection and slot filling, that extends JointBERT+CRF \cite{abs-1902-10909} with an intent-slot attention layer to explicitly convey the intent context information via the ``soft'' intent label embedding into slot filling. Our contributions are summarized as follows:

\begin{itemize}[leftmargin=*]
    \item We introduce the first public intent detection and slot filling dataset---named \textbf{PhoATIS}---for Vietnamese.
    \item We propose a joint model for intent detection and slot filling. Experimental results on our dataset show that: (i) our proposed model does significantly better than JointBERT+CRF; (ii) our attention mechanism is more effective than the previous ones \cite{goo2018slot,li-etal-2018-self,e-etal-2019-novel,qin-etal-2019-stack,zhang2019joint}; and (iii) automatic Vietnamese word segmentation and pre-trained monolingual language model are less effective for these Vietnamese intent detection and slot filling tasks than for other Vietnamese NLP tasks \cite{phobert,vitext2sql,PhoNER_COVID19}. 
    
    \item We publicly release our dataset and our model implementation for research or educational purpose. We hope that our dataset and model can serve as a starting point for future Vietnamese SLU research and applications.
\end{itemize}

\section{Related Work}\label{sec:related}

In addition to ATIS, SNIPS \cite{coucke2018snips} is also commonly used for intent detection and slot filling. However, recent performance scores reported on SNIPS are almost perfect \cite{Weld2021ASO,qin-etal-2019-stack,wang2020encoding}, thus resulting in a less challenging dataset. Given the popularity of ATIS, there are efforts to translate it into other languages. In particular, Upadhyay et al. \cite{upadhyay2018almost} and Xu et al. \cite{xu2020end}  extend  ATIS to eight more languages across four different language families. See \cite{Weld2021ASO} for a summary of  other intent detection and slot filling datasets.


Early research on intent detection and slot filling tackle these two tasks independently, where intent detection and slot filling are formulated as utterance classification and sequence labeling problems, respectively \cite{TurDHH12,ravuri2015recurrent,MesnilHDB13,vu2016bi}. Recent studies have shown that jointly learning intent detection and slot filling produces significant performance improvements over independent models \cite{zhang-etal-2019-joint,Weld2021ASO}. Two joint training strategies are investigated in the literature \cite{louvan2020recent}. The first strategy is through parameter and hidden state sharing, employing a shared BiLSTM/BERT encoder and two separate decoders for intent detection and slot filling that are structured on top of the encoder \cite{abs-1902-10909,XuS13,liu2016attention,3060832.3061040,hakkani2016multi}. The second strategy extends the first one to model the relationship between slots and intent labels. In particular, several research works  \cite{goo2018slot,li-etal-2018-self,e-etal-2019-novel} present attention mechanisms to compute the correlation between a global intent context representation and each slot vector outputted by the encoder; while other research works  \cite{qin-etal-2019-stack,zhang2019joint} first learn an utterance representation (i.e. equivalent to the global intent context representation) through self-attention and concatenate this representation with each of the encoder's vector outputs, before feeding the concatenated vectors into a slot filling decoder. See \cite{Weld2021ASO} for an overview of  other  methods for intent detection and slot filling.

\section{Our PhoATIS Dataset}


Our dataset construction process includes three manual phases. The first phase is to create a raw natural Vietnamese utterance set that is translated based on the ATIS dataset \cite{price-1990-evaluation}. The second phase is to project intent and slot annotations from ATIS to its Vietnamese-translated version. The last one is to fix inconsistencies among projected annotations.

\medskip
\noindent\underline{\textbf{First phase of translation:}}\ \ 
We manually translate all 5871 English utterances from ATIS into Vietnamese, including 4478, 500 and 893 utterances for training, validation and test, respectively \cite{abs-1902-10909,qin-etal-2019-stack}. The translation work is done by one NLP researcher and two research engineers with good English proficiency (IELTS 7.0+). We randomly split those English utterances into two non-overlapping and equal subsets. Every utterance from a subset is first translated by one engineer and then cross-checked and corrected by the second engineer; after that, the NLP researcher verifies each translated utterance and makes further revisions if needed. Here, there is a discussion session to finalize the best-translated version for each complicated case.

\textbf{NOTE} that unlike the multilingual ATIS datasets \cite{upadhyay2018almost,xu2020end} where the original slot values of American-specific entities in English are kept intact when translating into other languages; during our translation phase, we require adaptive modifications to make the translated utterances reflect real-world scenarios in the context of airline booking in Vietnam. In particular, there are 9336 slot values of American locations (e.g. airports, cities, and the like) and other American-popular entities (e.g. ticket codes, airlines, and the like); and the translation process replaces 8837/9336 slot values with their counterparts in Vietnam and all over the world. 
We also require to preserve spoken modalities (e.g. disfluency, word repetition and collocation) as much as possible to obtain a translated dataset that is correct, natural and similar to the real-world scenarios in Vietnam.

\medskip
\noindent\underline{\textbf{Second phase of annotation projection:}}\ \ 
We manually project the intent and slot annotations from the original ATIS dataset in English onto our Vietnamese-translated version. In this phase, each utterance in our Vietnamese dataset would have the same intent and slot label types as its corresponding English utterance. This annotation projection process is performed independently by the two research engineers. We again divide the dataset into 2 non-overlapping and equal subsets in which each subset is then annotation-projected by one engineer. This is a non-trivial annotation task because slot values and word orders are different between English utterances and their Vietnamese counterparts. After that, cross-checking is performed to ensure that there are no projection mistakes.

\begin{table}[!t]
     \centering    
     \caption{Statistics of our Vietnamese dataset PhoATIS with 28 intent labels and 82 slot types.}
    \begin{tabular}{l|l|l|l| l}
     \hline
         \textbf{Statistic} & \textbf{Train} & \textbf{Valid.} & \textbf{Test} & \textbf{All}  \\ \hline
         \# Utterances  & 4478 & 500 & 893 & 5871 \\ \hline
         \# Slots & 14859  & 1713  & 2842  & 19414  \\ \hline
     \end{tabular}
     \label{tab:stat}
\end{table}

\medskip
\noindent\underline{\textbf{Third phase of fixing inconsistencies:}}\ \  
Previous works also point out that there are errors in ATIS reference labels \cite{bechet2018atis, niu-penn-2019-rationally}. For example, all occurrences of the word token ``noon'' are labeled with the slot label type ``time''  in the training set, however, they are annotated with the type ``period\_of\_day'' in the validation set despite having similar contexts (e.g. ``noon'' in the training utterance ``which northwest flights stop in denver before \textbf{noon}'' and ``noon'' in the validation utterance ``please list only the flights from cleveland to dallas that leave before \textbf{noon}'' have labels ``time'' and ``period\_of\_day'', respectively). 
Regarding intent detection, there are also inconsistencies, e.g. the utterance ``what is the lowest fare from denver to atlanta'' annotated with the intent label ``airfare'' while the utterance ``show me the lowest price from dallas to baltimore'' annotated with the intent ``flight''. 

For our Vietnamese dataset, we also find similar inconsistent labels in both intent detection and slot filling when projecting the annotations in the previous second phase. We thus host another discussion session to refine annotation guidelines for handling annotation inconsistencies in our dataset. 
Then, based on the refined guidelines, we revisit each annotated Vietnamese utterance to make further corrections if needed,\footnote{Compared to the Vietnamese dataset outputted from the second phase, there are 146 changes in intent labels and 91 changes in slot annotations, across 198 utterances.} producing a final Vietnamese dataset of 5871 gold annotated utterances with 28 intent labels and 82 slot types.  Statistics of our dataset are shown in Table \ref{tab:stat}.

\begin{figure}[!t]
\centering
\includegraphics[width=7.75cm]{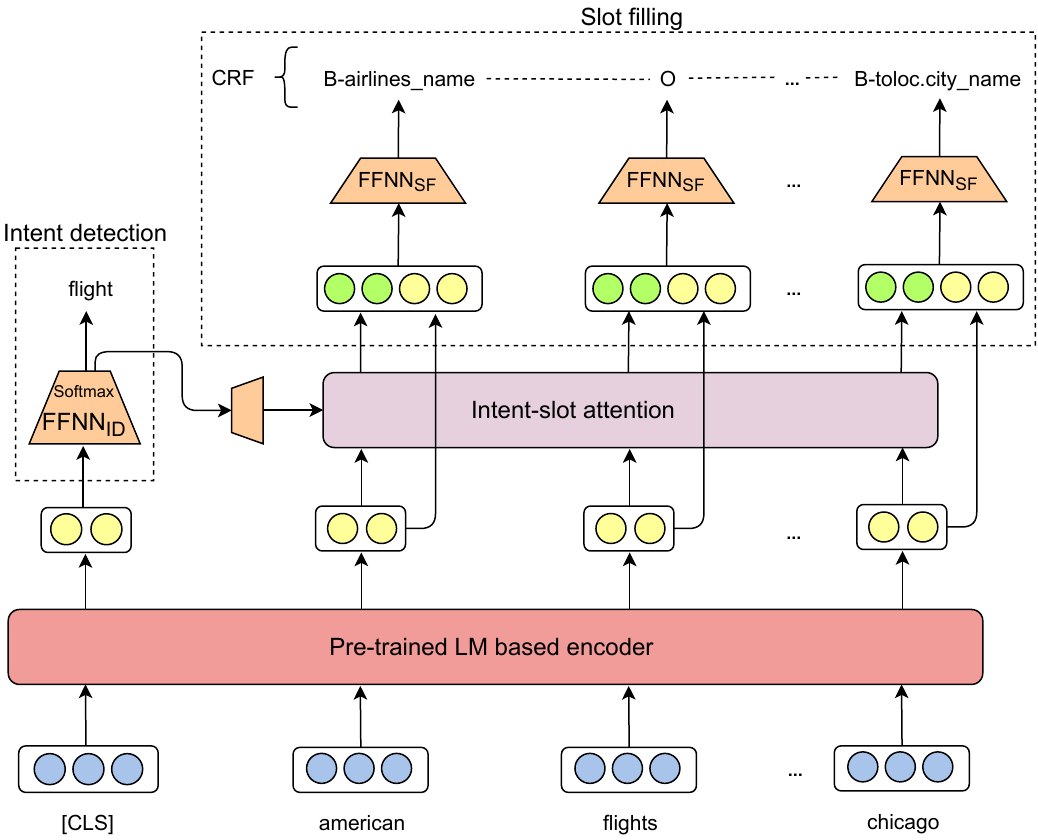}
\caption{Illustration of our proposed model JointIDSF.}
\label{fig:architecture}
\end{figure}

\section{Our Model} \label{sec:model}

Figure \ref{fig:architecture} illustrates the architecture of our  joint model---named \textbf{JointIDSF}---that consists of four layers including: an encoding layer (i.e. encoder), an intermediate intent-slot attention layer and two decoding layers of intent detection and slot filling. 

\medskip
\noindent\underline{\textbf{Encoding layer:}}\ \ 
Given an utterance consisting of $n$ tokens $w_1$, $w_2$, \dots, $w_n$, we insert a special classification token of ``[CLS]'' at the front of the utterance, resulting in an input utterance of $n+1$ tokens  $w_0$, $w_1$, $w_2$, \dots, $w_n$  for the encoding layer (here, $w_0$ is ``[CLS]''). The encoding layer employs a pre-trained Transformer-based language model (LM), e.g. BERT \cite{devlin-etal-2019-bert}, to produce contextualized latent feature embeddings $\mathbf{c}_{i} \in \mathbb{R}^{d_e}$ each representing the $i^{th}$ token $w_i$: 

\begin{equation}
\mathbf{c}_{i} = \mathrm{PretrainedLM}\big({w}_{0:n}, i\big)
\end{equation}

Here, $d_e$ is the embedding size of the encoder's contextualized embedding outputs. 

\medskip
\noindent\underline{\textbf{Intent detection layer:}}\ \ 
Following a common strategy when fine-tuning pre-trained LMs for a sequence classification task \cite{devlin-etal-2019-bert}, the intent detection layer is a linear prediction layer that is appended on top of the contextualized embedding $\mathbf{c}_{0}$ of the classification token ``[CLS]''. In particular, the intent detection layer feed $\mathbf{c}_{0}$ into a single-layer feed-forward network (FFNN\textsubscript{ID}) followed by a $\mathsf{softmax}$ predictor for intent prediction: 

\begin{equation}
\mathbf{p}  = \mathsf{softmax}\big(\mathrm{FFNN\textsubscript{ID}}\big(\mathbf{c}_{0}\big)\big)
\end{equation}

\noindent where the output layer size of FFNN\textsubscript{ID} is  $k$ being the total number of intent labels. Here, we formulate the prediction task as a multi-class classification problem.  Based on the probability vector $\mathbf{p} \in \mathbb{R}^{k}$, a cross-entropy objective loss {$\mathcal{L}_{\text{ID}}$} is calculated for intent classification during training. 

\medskip
\noindent\underline{\textbf{Intent-slot attention layer:}}\ \ 
We introduce an attention mechanism to align the importance of the intent information with each of the original utterance's tokens. In particular, the intent-slot attention layer takes the outputs from the encoding layer and the intent detection layer to produce intent-specific vectors that are then used as part of the input for the slot filling layer.   
Formally, this attention layer first creates a ``soft'' intent label embedding  $\mathbf{w} \in \mathbb{R}^{d_e}$ by multiplying a label weight matrix $\mathbf{W}\in \mathbb{R}^{d_e\times k}$ with the probability vector $\mathbf{p} \in \mathbb{R}^{k}$. Then it uses the intent label embedding $\mathbf{w}$ and the  contextualized embeddings  $\mathbf{c}_{i}$    to generate the intent-specific vectors $\mathbf{s}_{i}$ ($i \in \{1, 2, ..., n\}$) as follows:

 \begin{eqnarray}
 \mathbf{w} &=& \mathbf{W}\mathbf{p} \label{eqn:3} \\
 \alpha_i &=& \dfrac{\mathsf{exp}(\mathbf{w}^\top\mathbf{c}_i)}{\sum_{j=1}^n\mathsf{exp}(\mathbf{w}^\top\mathbf{c}_j)} \\
 \mathbf{s}_i &=& \alpha_i\mathbf{w} \label{eqn:5}
 \end{eqnarray}

\medskip
\noindent\underline{\textbf{Slot filling layer:}}\ \  
The slot filling layer formulates the slot filling task as a BIO-based sequence labeling problem. First, it creates a sequence of vectors  $\mathbf{v}_{1:n}$ in which each $\mathbf{v}_{i}$ is resulted in by concatenating the intent-specific vector $\mathbf{s}_{i}$  and the corresponding  contextualized embedding $\mathbf{c}_{i}$: 

\begin{equation}
\mathbf{v}_{i}  = \mathbf{s}_{i} \circ  \mathbf{c}_{i} \label{eqn:vi}
\end{equation}

 It then passes each vector $\mathbf{v}_{i}$ into another FFNN (FFNN\textsubscript{SF}):

\begin{equation}
 \mathbf{h}_{i} = \mathrm{FFNN\textsubscript{SF}}\big(\mathbf{v}_{i}\big)
\end{equation}

\noindent where the output layer size of FFNN\textsubscript{SF} is the number of BIO-based slot  types. 
Lastly, the slot filling layer feeds the output vectors $\mathbf{h}_{i}$ into a linear-chain CRF predictor  \cite{Lafferty:2001} for slot type prediction. A cross-entropy loss {$\mathcal{L}_{\text{SF}}$} is computed for slot filling during training while the Viterbi algorithm is used for inference.

\medskip
\noindent\underline{\textbf{Joint training:}}\ \ 
The final training objective loss \textbf{$\mathcal{L}$} of our joint model JointIDSF is the weighted sum of the intent detection  loss {$\mathcal{L}_{\text{ID}}$} and  the slot filling  loss {$\mathcal{L}_{\text{SF}}$}:

\begin{equation}
\textbf{$\mathcal{L}$} = \lambda\mathcal{L}_{\text{ID}} + (1 - \lambda)\mathcal{L}_{\text{SF}}
\end{equation}

\noindent where the hyper-parameter $\lambda$ is a mixture weight: $0 < \lambda < 1$.

\medskip
\noindent\underline{\textbf{Discussion:}}\ \ Our  JointIDSF  can be viewed as an extension of the recent state-of-the-art JointBERT+CRF model \cite{abs-1902-10909}, where we introduce the intent-slot attention layer to explicitly incorporate intent context information into slot filling. In particular,  without  the intent-slot attention layer,  Equation \ref{eqn:vi} would become $\mathbf{v}_{i}  =  \mathbf{c}_{i}$, and our model thus reduces to JointBERT+CRF.

Furthermore, our intent-slot attention layer is different from previous attention mechanisms \cite{goo2018slot,li-etal-2018-self,e-etal-2019-novel} in two important aspects: (i) we propose to use the intent label representation instead of the  ``[CLS]''-based utterance context representation (i.e. using  $\mathbf{w} =\mathbf{W}\mathbf{p}$ in Equation \ref{eqn:3} instead of $\mathbf{w} = \mathbf{c}_{0}$); and (ii)  our intent-slot attention layer's outputs are the scalar multiplication between the attention weights and the intent label representation  instead of the slot vector representations (i.e. using $\mathbf{s}_i = \alpha_i\mathbf{w}$ in Equation \ref{eqn:5} instead of $\mathbf{s}_i = \alpha_i\mathbf{c}_i$). In addition, using $\mathbf{v}_{i}  = \mathbf{c}_{0} \circ  \mathbf{c}_{i}$ (instead of $\mathbf{v}_{i}  = \mathbf{s}_{i} \circ  \mathbf{c}_{i}$ in Equation \ref{eqn:vi}) is equivalent to the approach  used in  \cite{qin-etal-2019-stack,zhang2019joint}.   Our ablation study results in Section \ref{ssec:main} show the effectiveness of our attention mechanism.

\section{Experiments}
\subsection{Experimental setup}

Note that the utterances in our PhoATIS dataset are annotated at the syllable level (as when written, the white space is both used to mark word boundaries as well as to separate syllables that constitute words). To obtain a word-level variant of the dataset, we perform automatic Vietnamese word segmentation by employing RDRSegmenter \cite{nguyen-etal-2018-fast} from VnCoreNLP \cite{vu-etal-2018-vncorenlp}. For example, a 4-syllable written text ``sân bay Nội Bài'' (Noi Bai airport) is word-segmented into 2-word text ``sân\_bay\textsubscript{airport} Nội\_Bài\textsubscript{Noi\_Bai}''. Here, the outputs of automatic Vietnamese word segmentation do not affect the span boundaries of slot annotations.

We conduct experiments on our dataset to study: (i) a quantitative
comparison between our model JointIDSF and the baseline JointBERT+CRF, (ii) the influence of Vietnamese word segmentation  (here, input utterances can be formed in either syllable or word level), and (iii) the usefulness of pre-trained language model-based encoders. Here, we employ XLM-R \cite{conneau-etal-2020-unsupervised} and PhoBERT \cite{phobert}---two recent state-of-the-art pre-trained language models that support Vietnamese---as the encoders. XLM-R, a multilingual variant of RoBERTa \cite{liu2019roberta}, is pre-trained on a 2.5TB multilingual dataset that contains 137GB of syllable-level Vietnamese texts. PhoBERT, a monolingual variant of RoBERTa for Vietnamese, is pre-trained on 20GB of word-level texts.

For both JointIDSF and JointBERT+CRF, we employ the AdamW optimizer \cite{loshchilov2018decoupled} and set the batch size to 32. We also perform grid search on the validation set to select their optimal hyper-parameters with the Adam initial learning rate in \{1e-5, 2e-5, 3e-5, 4e-5, 5e-5\} and the mixture weight $\lambda$ in \{0.05, 0.1, 0.15, ..., 0.9, 0.95\}.  
We train for 50 epochs, and calculate the average score of the intent accuracy for intent detection and the $F_1$-score (in \%) for slot filling after each training epoch on the validation set. We select the model checkpoint that obtains the highest average score over the validation set to apply to the test set. Note that our JointIDSF implementation initializes its encoder by a trained JointBERT+CRF's encoder. All our reported results are the average over 5 runs with 5 different random seeds.

\subsection{Main results}\label{ssec:main}
Table~\ref{tab:main_results} reports the test set results of our JointIDSF and the baseline JointBERT+CRF, employing standard evaluation metrics of the intent accuracy for intent detection, the $F_1$-score  for slot filling and the overall sentence accuracy \cite{louvan2020recent,Weld2021ASO}. The results are categorized into two comparable settings of using the syllable-level dataset and its automatically word-segmented variant for training and evaluation, associated with  the encoders XLM-R and PhoBERT, respectively. In each setting, we find that JointIDSF significantly outperforms JointBERT+CRF. Here, the highest improvements  are accounted for the sentence accuracy (i.e., 85.39\% $\rightarrow$ 86.17\% and 85.55\% $\rightarrow$ 86.25\%), thus showing that our intent-slot attention layer helps better capture correlations between intent labels and slots in the same utterances. We also find that the performances of word-level models are higher, but not significantly, than their syllable-level counterparts. Thus, automatic Vietnamese word segmentation and the pre-trained monolingual language model PhoBERT are less effective for these  Vietnamese intent detection and slot filling tasks than for other Vietnamese NLP tasks  \cite{phobert,vitext2sql,PhoNER_COVID19}. It is probably because the utterances in our dataset are domain-specific and medium-length ones with an average  length of 15 word tokens.

\begin{table}[!t]
    \centering
    \caption{Results on the test set.  ``Intent Acc.'' and ``Sent. Acc.'' denote intent detection accuracy and sentence accuracy, respectively. Each score improvement  over  JointBERT+CRF with the same encoder is statistically significant with p-value $< 0.05$ (except 97.56 vs 97.42 w.r.t. intent  accuracy).}
    \def\arraystretch{1.1}
    \setlength{\tabcolsep}{0.3em}
    \resizebox{8cm}{!}{
    \begin{tabular}{ll|l|c|c|c}
        \hline
         & \textbf{Model} & \textbf{Encoder}  & \textbf{Intent Acc.} & \textbf{Slot F1} & \textbf{Sent. Acc.}\\
        \hline
       \multirow{2}{*}{\rotatebox[origin=c]{90}{\textbf{Syll.}}} & JointBERT+CRF & XLM-R  & 97.42 & 94.62 & 85.39  \\
        & Our JointIDSF & XLM-R  & \textbf{97.56} & \textbf{94.95} & \textbf{86.17}  \\ 
        \hline
        \multirow{2}{*}{\rotatebox[origin=c]{90}{\textbf{Word}}}  & JointBERT+CRF & PhoBERT  & 97.40 & 94.75 &	85.55	 \\
        & Our JointIDSF & PhoBERT  & \textbf{97.62} & \textbf{94.98} & \textbf{86.25} \\
        \hline
    \end{tabular}
    }   
    \label{tab:main_results}
\end{table}

\begin{table}[!t]
    \centering
     \caption{Ablation study results on the validation set. Recall that JointBERT+CRF is a simplified variant of JointIDSF when using $\mathbf{v}_{i}  =  \mathbf{c}_{i}$ in Equation \ref{eqn:vi}. Each score difference between our full model JointIDSF and its ablated one is significant with p-value $< 0.05$ (except 98.45 vs. 98.35 w.r.t. intent  accuracy). } 
    \def\arraystretch{1.1}
    \setlength{\tabcolsep}{0.3em}
    \resizebox{8cm}{!}{
    \begin{tabular}{l|c|c|c}
        \hline
        \textbf{Model} & \textbf{Intent Acc.} & \textbf{Slot F1} & \textbf{Sent. Acc.}\\
        \hline
        Our JointIDSF\textsubscript{PhoBERT encoder}  & \textbf{98.45}	& \textbf{97.03} & \textbf{89.55} \\
        \hline
        \ \ \ \ (i) $\mathbf{w} = \mathbf{c}_{0}$ in Eq. \ref{eqn:3} & 98.05	& 96.62	& 88.30 \\
        \ \ \ \ (ii) $\mathbf{s}_i = \alpha_i\mathbf{c}_i$ in Eq. \ref{eqn:5} & 98.10	& 96.67 & 88.55 \\
        \ \ \ \ (iii) $\mathbf{v}_{i}  = \mathbf{c}_{0} \circ  \mathbf{c}_{i}$ in Eq. \ref{eqn:vi} & 98.35	& 96.78 & 88.85 \\
        \hdashline
        \ \ \ \ JointBERT+CRF\textsubscript{PhoBERT encoder} & 98.20 & 96.54 & 88.15 \\
        \hline 
    \end{tabular}
    }
   
    \label{tab:abalation}
\end{table}

We perform an ablation study to understand the model influences on the validation set using the word-level setup  (here, using the syllable-level setup results in similar findings). In particular, as discussed in Section \ref{sec:model}, we investigate the following factors: (i) using  the ``[CLS]''-based  utterance context representation $\mathbf{c}_{0}$ instead of the intent label representation (i.e. using  $\mathbf{w} = \mathbf{c}_{0}$ in Equation \ref{eqn:3}), (ii) using the scalar multiplication between the attention weights and  the slot vector representations $\mathbf{c}_{i}$ instead of the intent label representation  (i.e. using $\mathbf{s}_i = \alpha_i\mathbf{c}_i$ in Equation \ref{eqn:5}), and (iii) concatenating the utterance context representation $\mathbf{c}_{0}$ instead of the attention layer's output vectors $\mathbf{s}_i$ to all the slot vector representations (i.e. using $\mathbf{v}_{i}  = \mathbf{c}_{0} \circ  \mathbf{c}_{i}$ in Equation \ref{eqn:vi}). Table \ref{tab:abalation} shows that all these  factors  degrade the performance of our full model, clearly showing the more effectiveness of our attention mechanism compared to the previous ones \cite{goo2018slot,li-etal-2018-self,e-etal-2019-novel,qin-etal-2019-stack,zhang2019joint}.

\subsection{Error analysis}

We also provide an illustration example to compare prediction outputs of JointIDSF and JointBERT+CRF w.r.t. the validation utterance ``chuyến bay nào rời  sân\_bay  vân\_đồn đến  côn\_đảo  và hạ\_cánh lúc  10 giờ tối'' (what flights leave Van Don airport for Con Dao and arrive at 10 pm). Both JointIDSF and JointBERT+CRF predict the intent ``flight'' for this utterance correctly. In addition, our JointIDSF correctly recognizes  ``sân\_bay\textsubscript{airport} vân\_đồn\textsubscript{Van\_Don}'' as an airport of departure location, that is tagged with the slot type ``fromloc.airport\_name''. However, JointBERT+CRF is failed,  tagging ``sân\_bay\textsubscript{airport}'' with the slot type ``fromloc.city\_name''---the name of a departure city, and not recognizing  ``vân\_đồn\textsubscript{Van\_Don}'' as a part of any slot. Another example is ``cho tôi danh\_sách các chuyến bay vào ngày 27 tháng 12 từ đài\_bắc đến singapore và giá vé tương\_ứng'' (give me a list of flights on 27 December from Taipei to Singapore and their corresponding airfare). Both  JointIDSF and JointBERT+CRF predict  slots correctly. However, while our JointIDSF produces a correct intent label ``airfare\#flight'', JointBERT+CRF produces an incorrect intent label of ``flight''.

\begin{table}[!t]
    \centering
    \caption{Counts for error types on the validation set of our JointIDSF\textsubscript{PhoBERT encoder} (average over the 5 different runs).}
    \def\arraystretch{1.1}
    \setlength{\tabcolsep}{0.3em}
    \resizebox{7.75cm}{!}{
    \begin{tabular}{p{8cm} | c}
        \hline
        Definition & \#errors\\
        \hline
         Wrong Intent (\textbf{WI}): Predicted intent label is not the gold-annotated  one. & 8 \\
        \hline
         Missing Slot (\textbf{MS}): A gold slot's span is not entirely or partly recognized. & 5\\
        \hline
         Spurious Slot (\textbf{SS}): A predicted slot matches a gold  O label. & 10\\
        \hline
         Wrong Boundary (\textbf{WB}): A predicted slot's span is partly overlapped with a gold slot's span, while the predicted slot's label type is the gold slot's. & 14 \\
        \hline
         Wrong Label (\textbf{WL}): The predicted slot has exact span boundary while having incorrect slot label. & 29 \\
        \hline
    \end{tabular}
    }
    \label{tab:error}
\end{table}

To understand the source of errors, we perform an error analysis using the best performing model JointIDSF\textsubscript{PhoBERT encoder} on the word-level validation set. We categorize all error cases into different categories of WI, MS, SS, WB and WL, as listed in Table \ref{tab:error}. There are 8 errors counted for the WI category, and most of them are induced by the multi-intent labels since the model is likely to predict the most clearly manifested or first appeared intent. For example, the model predicts an intent label of ``airfare'' instead of the gold one ``airfare\#flight\_time'' for the utterance ``cho tôi biết  chi\_phí  và  thời\_gian của  các chuyến bay  từ  phú\_quốc đến  cam\_ranh'' (show me the cost and time for flights from Phu Quoc to Cam Ranh).  
There are 5 and 10 errors counted for the error categories MS and SS, respectively. For these two types of errors, the model is often ambiguous about the slot types that rarely appear in the training set such as ``connect'', ``airport\_code'' and the like. 
The WB error category has  14 error cases that are related to multi-word spanned slots. 
The remaining 29 error cases are accounted for the WL error category. These cases often are induced by ambiguities between which the ``departure'' part is and which the ``arrival'' part is in an utterance since  many utterances do not have an explicit context. For example, given the utterance ``tôi  cần  đến  phú\_quốc vào  tối  thứ\_tư  từ  đà\_lạt'' (I need to go to Phu Quoc on Wednesday's night from Da Lat), it is relatively ambiguous to determine whether the phrase ``tối thứ\_tư'' (Wednesday's night) refers to as an arrival time or a departure time without a clearer context.

\section{Conclusion}
In this paper, we have presented the {first} public dataset for Vietnamese intent detection and slot filling. In addition, we also have proposed an effective model, namely JointIDSF, for jointly learning intent detection and slot filling. In particular, JointIDSF extends the recent state-of-the-art JointBERT+CRF \cite{abs-1902-10909} by introducing the intent-slot attention layer to  incorporate intent context information into slot filling explicitly. We empirically conduct experiments and perform a detailed error analysis on our dataset, and show that: JointIDSF  significantly outperforms JointBERT+CRF and our attention mechanism is more effective than the previous ones \cite{goo2018slot,li-etal-2018-self,e-etal-2019-novel,qin-etal-2019-stack,zhang2019joint}. 
We hope that the public release of our dataset and JointIDSF implementation can serve as the starting point for further research and applications in Vietnamese spoken and natural language understanding.

\bibliographystyle{IEEEtran}
\bibliography{mybib}

\end{document}